\documentclass[runningheads]{llncs}
\usepackage{graphicx}
\usepackage{amsmath,amssymb} 
\usepackage{color}

\usepackage{tabularx}
\usepackage{algorithm}
\usepackage{algpseudocode}
\usepackage{amsmath}
\usepackage{amssymb}
\usepackage{diagbox}

\usepackage{pifont}
\newcommand{\cmark}{\ding{51}}%
\newcommand{\xmark}{\ding{55}}%

\begin{document}
\pagestyle{headings}
\mainmatter

\def\ACCV20SubNumber{826}  

\title{Semantics through Time: Semi-supervised Segmentation of Aerial Videos with \\ Iterative Label Propagation} 
\titlerunning{Semantics through Time}
%
\author{Alina Marcu\inst{1, 2} \and
Vlad Licaret\inst{1} \and Dragos Costea\inst{1,2} \and Marius Leordeanu\inst{1,2}}
\authorrunning{A. Marcu et al.}
%
\institute{University Politehnica of Bucharest, 313 Splaiul Independentei, Bucharest \and
Institute of Mathematics of the Romanian Academy, 21 Calea Grivitei, Bucharest\\
\email{\{alina.marcu,vlad.licaret,dragos.costea,marius.leordeanu\}@upb.ro}}

\maketitle

\begin{abstract}
Semantic segmentation is a crucial task for robot navigation and safety. However, current supervised methods require a large amount of pixelwise annotations to yield accurate results. Labeling is a tedious and time consuming process that has hampered  progress  in  low  altitude  UAV applications. This paper makes
an important step towards automatic annotation by introducing \textit{SegProp}, a novel iterative flow-based  method, with a direct connection to spectral clustering in space and time, to propagate the semantic labels to frames that lack human annotations. The labels are further used in semi-supervised learning scenarios. Motivated by the lack of a large video aerial dataset, we also introduce \textit{Ruralscapes}, a new dataset with high resolution (4K) images and manually-annotated dense labels every 50 frames - the largest of its kind, to the best of our knowledge. 
Our novel SegProp automatically annotates the remaining unlabeled 98\% of frames with an accuracy exceeding $90\%$ (F-measure), significantly outperforming other state-of-the-art label propagation methods. Moreover, when integrating other methods as modules inside SegProp's iterative label propagation loop, we achieve a significant boost over the baseline labels. Finally, we test SegProp in a full semi-supervised setting: we train several state-of-the-art deep neural networks
on the SegProp-automatically-labeled training frames and test them on completely novel videos. We convincingly demonstrate, every time, a significant improvement over the supervised scenario.
\end{abstract}


\section{Introduction}


While ground vehicles are restricted to movements in 2D, aerial robots are free to navigate in three dimensions. This allows them to capture images of objects from a wide range of scales and angles, with richer views than the ones available in datasets collected on the ground. Unfortunately, this unconstrained movement imposes significant challenges for accurate semantic segmentation, mostly due to the aforementioned variation in object scale and viewpoint. 
Classic semantic segmentation approaches are focused on ground scenes. More recent work tackled imagery from the limited viewpoints of specialized scenes, such as ground-views of urban environments (from vehicles) and direct overhead views (from orbital satellites). Nevertheless, recent advances in aerial robotics allows us to capture previously unexplored viewpoints and diverse environments more easily. Given the current state of technology, in order to evaluate the performance of autonomous systems, the human component is considered a reference. However, the manual segmentation annotations in supervised learning is a laborious and time consuming process. In the context of video segmentation, it is impractical to manually label each frame independently, especially considering there is relatively little change from one to the next. In this context, the ability to perform automatic annotation would be extremely valuable.

\begin{figure}[!t]
\centering
\includegraphics[scale=0.3,keepaspectratio]{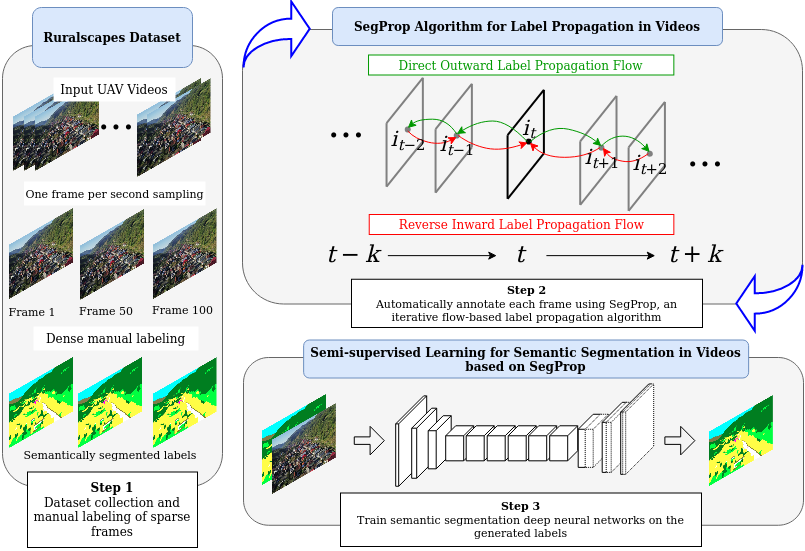}
\caption{\label{fig:overview} \textbf{SegProp}: our method for automatic propagation of semantic labels in the context of semi-supervised segmentation in aerial videos. \textbf{Step 1.} First, we sample the UAV videos, at regular intervals (e.g. one or two frames per second). The resulting frames are then manually labeled. \textbf{Step 2.} We automatically propagate labels to the remaining unlabeled frames using our SegProp algorithm - based on class voting, at the pixel level, according to inward and outward label propagation flows between the current frame and an annotated frame. The propagation flows could be based on optical flow, homography transformation or other propagation method, as shown in experiments. SegProp propagates iteratively the segmentation class voting until convergence, improving performance over iterations.
\textbf{Step 3.} We then mix all the generated annotations with the ground truth manual labels to train powerful deep networks for semantic segmentation and significantly improve performance in unseen videos.}
\end{figure}

\textbf{SegProp - automated semantic propagation in videos.}
In this paper we present SegProp (Sec. \ref{sec:segprop}), 
an iterative flow-based  method to propagate, through space and time, the semantic segmentation labels to video frames that lack human annotations. SegProp propagates labels in an iterative fashion, forward and backward in time from annotated frames,
by looping several times through the video and accumulating class votes at each iteration. At convergence the majority class wins. From a theoretical point of view, SegProp relates to spectral MAP labeling in graphical models and has convergence and improvement guarantees (Sec.~\ref{sec:math_interp}). In practice, we demonstrate the effectiveness of SegProp in several ways (Sec.~\ref{sec:experiments}). First, we show that SegProp is able to propagate labels to unlabeled frames with an accuracy that outperforms competition by a significant margin (Sec.~\ref{sec:exp_comparisons}). Second, we show that other methods for label propagation could be immediately integrated as modules inside the SegProp propagation loop, with a significant boost in performance (Sec.~\ref{sec:exp_ablation}).
And third, we demonstrate SegProp's effectiveness in a semi-supervised learning scenario (Sec.~\ref{sec:exp_semi_sup}), 
in which several state-of-the-art deep networks for semantic segmentation are trained on the automatically annotated frames and tested on novel videos, with an important improvement over the supervised case. 

\textbf{Label propagation methods}. Our method is not limited to single-object maks propagation~\cite{le2016geodesic}. Methods that are most related to ours perform total scene, multi-class label propagation~\cite{galasso2012video}. One such method, also propagates labels between two frames, in the context of ground navigation and low resolution images (320 x 240)~\cite{budvytis2017large}. They employ an occlusion-aware algorithm coupled with an uncertainty estimation method. Their approach is less useful in our case, where we have very high resolution images (4k) at a high frame rate (50fps) and dense optical flow can be accurately computed. Earlier works, exploring the idea of propagating ground truth labels using an optical flow based method~\cite{mustikovela2016can}, have shown that it could be useful to treat pseudo-labels differently than the ground truth ones, during training. That idea builds upon other work that addresses occlusion errors~\cite{chen2010propagating}. More recent methods for automatic label propagation use a single human annotated frame and extend the label to nearby frames, such as it is the work of Zhu et al.~\cite{zhu2019improving}, with sate-of-the-art results on
Cityscapes and KITTI~\cite{geiger2013vision}. The main limitation of~\cite{zhu2019improving}
is that the performance drops drastically when increasing the number of propagated frames, as we confirm in our tests (Fig. \ref{fig:homography_results}). Budvytis et al.~\cite{budvytis2010label} use semi-supervised learning to improve the intermediate labels and Reza et al.~\cite{reza2017label} integrate depth and camera pose and formulate the problem as energy minimization in Conditional Random Fields. 

\textbf{Ruralscapes Dataset for Semantic Segmentation in UAV Videos.}
In this paper we also introduce \textit{Ruralscapes}, the largest high resolution video dataset (20 high quality 4K videos) for aerial semantic segmentation, taken in flight over rural areas in Europe (Sec. \ref{sec:ruralscapes}). We manually annotate a relatively small subset ($2\%$) of frames in a video and use SegProp, our novel iterative label propagation algorithm, to automatically annotate the whole sequence. Given a start and an end frame of a video sequence, SegProp finds pixelwise correspondences between labeled and unlabeled frames, to assign a class for each pixel in the video based on an iterative class voting procedure. In this way we generate large amounts of labeled data (over 50k densely segmented frames) to use in semi-supervised training deep neural networks and show that training on the automatically generated labels, boosts the performance at test time significantly.
Our pipeline can be divided into three steps (see Figure \ref{fig:overview}). The first and most important is the data labeling step. We leverage the advantages of high quality 4K aerial videos, such as small frame-to-frame changes (50 frames per second) and manually annotate a relatively small fraction of frames, sampled at 1 frame per second.
We then automatically generate labels for each intermediate frame, between two labeled ones using SegProp, our proposed algorithm (Section~\ref{sec:segprop}), such that the whole video is labeled.
In our last step, we use the manually and automatically annotated frames together for semi-supervised training. 

\textbf{Datasets for semantic segmentation in video}. Since most work is focused on ground navigation, the largest datasets with real-world scenarios are ground-based. Earlier image-based segmentation datasets, such as Microsoft's COCO~\cite{lin2014microsoft}, contained rough labels, but the large number of images (123k) and classes (80), made it a very popular choice. 
Cityscapes~\cite{Cordts2016Cityscapes} was among the first large-scale dataset for ground-level semantic and instance segmentation. Year after year, the datasets increased in volume and task complexity, culminating with Apolloscape~\cite{huang2018apolloscape}, which is, to the best of our knowledge, the largest real ground-level dataset. Compared to its predecessors, it also includes longer video shots, not just snippets. It comprises of 74,555 annotated video frames. To help reduce the labeling effort, a depth and flow-based annotation tool is used.
Aeroscapes~\cite{nigam2018ensemble} is a UAV dataset that contains real-world videos and semantic annotations for each frame and it is closer to what we aim to achieve. Unfortunately, the size of the dataset is rather small, with video snippets ranging from 2 to 125 frames. The most similar dataset to ours is UAVid ~\cite{lyu2018uavid}. The dataset contains set of 4K UAV videos, that captures urban street scenes, with 300 images manually-labeled with 8 classes, compared to our dataset that has 60\% more frames, manually-labeled with 12 classes. 
Since labeling real-world data (especially video) is difficult, a common practice is to use synthetic videos from a simulated environment. Such examples are Playing for Benchmarks~\cite{richter2017playing}, for ground-level navigation and the recently released Mid-air~\cite{Fonder2019MidAir}, for low-altitude navigation. Mid-air has more than 420k training video frames. The diversity of the flight scenarios and classes is reduced - mostly mountain areas with roads - but the availability of multiple seasons and weather conditions is a plus.

\textbf{Main contributions: 1)} We present SegProp, an iterative semantic label propagation method in video, which outperforms the current state-of-the-art (Sec. \ref{sec:segprop}). \textbf{2)} We introduce Ruralscapes, the largest high-res (4K) video dataset for aerial semantic segmentation with 50,835 fully annotated frames and 12 semantic classes (Sec. \ref{sec:ruralscapes}). \textbf{3)} SegProp can be easily integrated with other label propagation methods and further improve their initial segmentation results (Sec. \ref{sec:exp_ablation}).
\textbf{4)} We test SegProp in semi-supervised learning scenarios and compare with state-of-the-art deep neural nets for semantic segmentation (Sec. \ref{sec:exp_semi_sup}).

\section{SegProp: Semantic Propagation through Time}
\label{sec:segprop}

We now present SegProp, our iterative, voting-based label propagation method (Alg. \ref{alg:labelInterpIterations}), which
takes advantage of the temporal coherence and symmetry present in videos. Before presenting the full method, we first show how labels are propagated between two labeled frames to the intermediate initially unlabeled ones (in one iteration).
Let $P_{k}$ be an intermediate (initially unlabeled) video frame between two (manually) labeled key frames $P_{i}$ and $P_{j}$. We first extract optical flow both forward $F_{i\rightarrow j}$ and backward $F_{j\rightarrow i}$ through time, between subsequent frames. Then, we use the dense pixel motion trajectories formed by the optical flow and map pixels from the annotated frames towards the unlabeled $P_{k}$. Since the optical flow mapping is not bijective (mapping from $P_{k}$ to $P_{j}$ could differ from mapping from $P_{j}$ to $P_{k}$), we take both forward and backward mappings into account. Thus, for each pixel in $P_{k}$ we have 4 correspondence maps that will vote for a certain class: two votes are collected based on the direct outward maps from $P_{k}$ to its nearby key labeled frames ($P_{k\rightarrow i}$,$P_{k\rightarrow j}$) and two are based on the reversed inward maps ($P_{i\rightarrow k}$, $P_{j\rightarrow k}$). Since motion errors are expected to increase with the length of the temporal distance between frames, we weigh these votes with exponential decay, decreasing with the distance (Alg. \ref{alg:labelInterp}). 

\textbf{Notation.} $P_k$ is a 3-dimensional segmentation map, of the same two dimensions as the frame, but with a third dimension corresponding to the class label. Thus, votes for a given class are accumulated on the channel corresponding to that specific class. With a slight notation abuse, by $P_{k\rightarrow i}$ we denote either the flow propagation directed from frame $k$ to frame $i$ as well as the class vote cast by the labeled frame (in this case, $i$) onto the unlabeled frame $k$ at the corresponding locations in $k$, according to the flow propagation map $P_{k\rightarrow i}$ .

\begin{algorithm}
\caption{Label propagation between labeled frames in one iteration}
\label{alg:labelInterp}
\begin{algorithmic}
    \State \textbf{Input:} Two labeled frames $P_i$ and $P_j$, optical flow maps $F_{i\rightarrow j}$ and $F_{j\rightarrow i}$. 
    \State \textbf{Output}: $P_k$, an intermediate, automatically labeled frame ($i < k < j$).
    \State 1) Compute 4 segmentations $P_{kn}$ ($n = \overline{1,4}$) by casting
    class votes ($P_{k\rightarrow i}$, $P_{k\rightarrow j}$) and ($P_{i\rightarrow k}$, $P_{j\rightarrow k}$) according to direct outward flow and reverse inward flow, respectively.
    \State 2) Gather weighted votes from all four $P_{kn}$ for each pixel $(x,y)$:
    \State \hspace{1cm} $p_k(x,y,:) = \sum_{n=1}^{4} w_n p_{kn}(x, y,:)$, 
    \State \hspace{1cm} where $w_n \propto e^{-\lambda * dist_n(k,q_n)}$
    and $dist_n(k, q_n)$ is the distance between frame $k$ and corresponding labeled frame $q_n \in {i,j}$. 
    Weights $w_n$ are normalized to sum to 1.
    \State 3) Compute final $P_k$ by class majority voting for each pixel.
\end{algorithmic}
\end{algorithm}

\textbf{Iterative SegProp Algorithm.} A similar label propagation procedure (as in Alg. \ref{alg:labelInterp}) could be repeated for several iterations (as in Alg. \ref{alg:labelInterpIterations}) by considering all frames labelled from previous iterations and cast votes between nearby ones. The intuition is that after the initial voting, we can establish better temporal coherence among neighbouring frames and improve consistency by iteratively propagating class votes between each other. The iterative SegProp (Alg. \ref{alg:labelInterpIterations}) results in better local consensus, with smoother and more accurate labels. In Sec. \ref{sec:math_interp} we also show that SegProp has interesting theoretical properties such as convergence to an improved segmentation objective score.

\textbf{Integrating other propagation methods into SegProp.}
We can use SegProp as a meta-procedure on top of other label propagation solutions (such as \cite{zhu2019improving} or homography-based propagation), resulting in further improvement of the initial results, as shown in our experiments (Sec. \ref{sec:experiments}). Segprop could in principle start from any initial solution (soft or hard) and then, at each iteration, replace or augment the class voting with votes from any other label propagation module. 

\begin{algorithm}[!t]
\caption{Iterative SegProp Algorithm for Label Propagation}
\label{alg:labelInterpIterations}
\begin{algorithmic}
    \State 1) For a given frame $k$, perform \textbf{Step 1} from \textbf{Algorithm 1} considering of $2f$ neighbouring frames, symmetrically spaced around $k$. \textit{ Optional: replace or augment the optical flow votes by another propagation procedure.}
    \State 2) Gather all available votes for each class, including the ones from frame $k$'s previous iteration:
        \State \hspace{1cm} $p_k(x,y,:) = \sum_{n=1}^{4f+1} w_n p_{kn}(x, y,:)$
    \State 3) Return to Step 1 and repeat several iterations, for all frames, until convergence.
    \State 4) Return final segmentation by class majority voting: $class_k(x,y) = max(p_k(x,y,:))$
\end{algorithmic}
\end{algorithm}

\textbf{Final segmentation space-time 3D filtering.} As final post-processing step 
we smooth out the segmentation noise as follows: we propagate $P_k$ along optical flow vectors for a number of steps, forward and backward through time, and concatenate the results into a local 3D spatiotemporal voting volume, one per class. We then apply a 3D (2D + time) Gaussian filter kernel to the 3D volume and obtain an average of the votes, one per each class, independently. Then we finally set hard per-pixel classes by class majority voting.


\subsection{Mathematical interpretation and properties of SegProp}
\label{sec:math_interp}

From a more theoretical point of view, SegProp can be seen as a Maximum A Posteriori (MAP) label inference
method for graphical models \cite{li1994markov}, strongly related to other, more classical iterative optimization techniques for labeling problems with pairwise terms, such as relaxation labeling \cite{hummel1983foundations},
deterministic and self annealing \cite{rangarajan2000self}, spectral MAP inference \cite{leordeanu2006efficient} and
iterative conditional modes \cite{besag1986statistical}.
Conceptually, we could think of the video as a graph of pixels in space and time, 
with a node for each pixel. Then, each node in the graph can get one class label out of several.
A multi-class segmentation solution at the entire video level, 
could be represented with a single vector $\mathbf{p}$, with N $\times$ C elements ($N$ - total number of nodes,
$C$ - number of possible classes per node). Thus, for a unique pixel $i$ in the video and
potential label $a$, we get a unique index $ia$. A final hard segmentation could then be expressed
as an indicator vector $\mathbf{p}$, such that $p_{ia}=1$ if pixel $i$ has class $a$ and $p_{ia}=0$, otherwise.

We consider the space-time 
graph edge structure as given by the flow based links between neighbouring 
pixels as presented in Algorithms 1 and 2. Thus, any two pixels $(i,j)$ connected through an optical flow link that vote for the same class (where pixel $i$ is from one frame and pixel $j$ from another) establish an undirected edge between them. 
These edges define the structure of the graph, with adjacency matrix $\mathbf{M}$.

Then, the weighted class voting can be expressed by correctly defining $\textbf{M}$, such that
$\mathbf{M}_{ia,jb} = w_{ij} = e^{-\lambda * dist(i,j)}$ if and only if $(i,j)$ are connected and class $a$ is the same
as class $b$ (class $a$ from frame of pixel $i$ can only vote for the same class in frame of pixel $j$).
One can then show that the iterative voting procedure (Alg. 2) can be written in simplified matrix form as:

\begin{equation}
\label{eq:SegProp_Power_Iteration}
      \mathbf{p}_{ia}^{(t+1)} \gets \sum_{j \in N_i} \mathbf{M}_{ia,ja} \mathbf{p}_{ja}^{(t)}
\end{equation}
      
Note that the sum above is exactly the accumulation of votes coming from the neighbouring frames (labeled at the previous iteration).
Also note that as the number of votes per node over all classes is constant (equal to the total number of votes), 
the vector $\mathbf{p}$ remains $L1$-normalized from one iteration to the next, 
both at the local level of nodes ($\sum_a \mathbf{p}_{ia} = N_{votes}$)
and overall $\sum_{ia} \mathbf{p}_{ia} = N_{pixels}*N_{votes}$. One could 
immediately observe that Equation \ref{eq:SegProp_Power_Iteration} above is the power iteration method
for computing the principal eigenvector of matrix $\mathbf{M}$, which must have positive elements since matrix
$M$ has positive elements, according to Perron-Frobenius theorem. It also means 
that the final solution of SegProp should, in principle, 
always converge to the same solution regardless of initialization as it depends only on $\mathbf{M}$,
which is defined by the propagation flow of labels (e.g. optical flow) and the initial manually labeled
frames. Note that those frames should never change their solution and never accumulate votes, a condition that can be easily enforced through
the way we set up $\mathbf{M}$. 

It is also well known that the principal eigenvector of $\mathbf{M}$ maximizes a segmentation score
$S_{L}(\mathbf{p}) = \mathbf{p}^{T} \mathbf{M} \mathbf{p}$, under L2-norm constraints on $\mathbf{p}$. In other words, SegProp should converge to $\mathbf{p}^{*} = \operatorname*{argmax}_p S(\mathbf{p})= \sum_{ia} \sum_{jb} \mathbf{M}_{ia,jb} \mathbf{p}_{ia} \mathbf{p}_{jb}$. Now, if we use the definition of our pairwise terms in $\mathbf{M}$ we can also show that:  $S_{L}(\mathbf{p}) = \sum_{j \in N_{i}} \mathbf{M}_{ia(i), ja(i)} \mathbf{p}_{ia(i)} \mathbf{p}_{ja(i)}=\sum_{i}N_{i}(a(i))$, where $N_{i}(a(i))$ are the number of neighbours of node $i$, which have the same label $a(i)$ as $i$. Thus, maximizing the segmentation score has a natural and intuitive meaning: we will find the segmentation $\mathbf{p}$ that encourages connected nodes to have the same label. 

\textbf{SegProp theoretical properties.} In summary, we expect SegProp to converge and maximize the quadratic soft-segmentation score with pairwise links $S_{L}(\mathbf{p}) = \mathbf{p}^{T} \mathbf{M} \mathbf{p}$, under L2-norm constraints on $\mathbf{p}$. 
It should do so regardless of the initialization as it only depends on $\mathbf{M}$, defined by the propagation flow and the initial manually labeled frames. Initialization should, however, affect the speed of convergence, as also observed in experiments (Sec. \ref{sec:exp_comparisons}). Since the segmentation $\mathbf{p}$ has constant L1 norm, we also expect it to converge to the stationary distribution of the random walk, as defined by the transition adjacency matrix $\mathbf{M}$. Thus, the solution, which is the principal eigenvector of $\mathbf{M}$, strongly relates SegProp to spectral clustering \cite{meila2001random} and spectral MAP inference \cite{leordeanu2006efficient}, a fact that could help us better understand its behaviour in practice.

\section{The Ruralscapes Dataset}
\label{sec:ruralscapes}


\textbf{Manual annotation tool.} In order to manually annotate the sampled frames, we designed a user-friendly tool that facilitates drawing the contour of objects (in the form of polygons). For each selected polygon we can assign one of the 12 available classes, which include background regions: forest, land, hill, sky, residential, road or water, and also foreground, countable objects: person, church, haystack, fence and car.
In the context of total scene segmentation, we assume that the image needs to be fully segmented (e.g., no 'other' class). While there are other annotation tools available (\cite{cocoannotator},~\cite{russell2008labelme}), ours has several novel convenient features for rapid annotation that go beyond simple polygonal annotation.

Our software is suited for high resolution images. Furthermore, it offers support for hybrid contour and point segmentation - the user can alternate between point-based and contour-based segmentation during a single polygon. The most time-saving feature is a 'send to back' functionality to copy the border from the already segmented class to the new one being drawn. This tool is mostly useful in cases when smaller objects are on top of bigger ones (such as cars on the road). Instead of delineating the area surrounding the car twice, one can firstly contour the road and on top that polygon, segment the car. 

\begin{figure}[!t]
\includegraphics[width=1.0\linewidth,keepaspectratio]{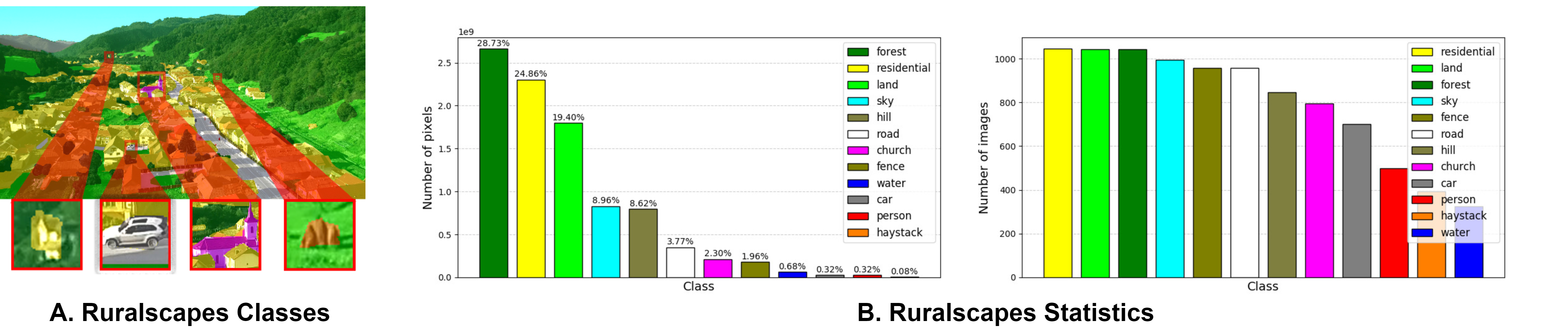}
\caption{\label{fig:ruralscapes} \textbf{A. Ruralscapes classes}. Labels overlaid over RGB image with detail magnification, offering a good level of detail. Ruralscapes also offers large variation in object scale. \textbf{B. Ruralscapes statistics}. (Left) Class pixels' distribution. Being a rural landscape, the dominant classes are buildings, land and forest. Due to high altitude, smaller classes such as haystack, car and person hold a very small percentage. (Right) Number of labeled images in which each class is present. 
}
\end{figure} 

\textbf{Dataset details and statistics.} We have collected 20 high quality 4K videos portraying rural areas. Ruralscapes comprises of various landscapes, different flying scenarios at multiple altitudes and objects across a wide span of scales. The video sequence length varies from 11 seconds up to 2 minutes and 45 seconds. The dataset consists of 17 minutes of drone flight, resulting in a total of 50,835 fully annotated frames with 12 classes. Of those, 1,047 were manually annotated, once every second. To the best of our knowledge, it is the largest dataset for semantic segmentation from real UAV videos.
The distribution of classes in terms of occupied area is shown in Figure~\ref{fig:ruralscapes} (B). Background classes such as forest, land and residential are dominant, while smaller ones such as person and haystack are at the opposite spectrum.

\textbf{Details regarding the annotation process.} Labels offer a good level of detail, but, due to the reduced spatial resolution of the small objects, accurate segmentation is difficult, as seen in the sample label from Figure~\ref{fig:ruralscapes} (A). Some classes, such as haystack, are very small by the nature of the dataset, others such as person, also feature close-ups. 
Manual labeling is a time consuming process. Based on the feedback received from the 21 volunteers that segmented the dataset, it took them on average 45 minutes to label an entire frame. This translates into 846 human hours needed to segment the manually labeled 1,047 frames.


\section{Experimental Analysis}
\label{sec:experiments}

We evaluate the performance of our proposed method and compare the results to the current state-of-the-art for label propagation~\cite{zhu2019improving}. We also train three widely adopted segmentation networks on the automatically generated segmentation labels and report the results compared to the baseline supervised training. 

\textbf{Dataset split.} The whole 20 densely labeled video sequences are divided into 13 training and 7 testing video subsets. We divided the dataset in such a way to be representative enough for the variability of different flying scenarios. We selected the videos in such a manner that the scenes were equally distributed between the training and testing sets. We never have the same exact scene in both train and test. When two videos have similar scenes (w.r.t classes, viewpoint and altitude), we select the longer video for train and the other for test (happened only once).
The 7 test videos ($\approx$29.61\% of the total frames from the dataset) have 311 manually-labeled frames (used for evaluation metrics) out of a total of 15,051 frames. The 13 training videos have 736 manually-labeled frames out of a total of 35,784 frames that we automatically annotate, starting from the initial manually labeled ones, using SegProp for the semi-supervised learning tests. 


For experimental purposes, we sample manually labeled frames every 2 seconds (every 100th frame, starting with the first) from the training set, and term this set \textit{TrainEven}. The remaining manually-labeled frames, the ones at odd seconds marks, form the 
\textit{TrainOdd} set, which are used, as explained later, to test the performance of label propagation on the training set itself, before semi-supervised deep learning and evaluation on the unseen test videos. We conducted a more detailed analysis of the influence of having larger temporal gaps between labeled frames over the segmentation performance, shown in Figure \ref{fig:homography_results} (B).

\subsection{Comparisons to other label propagation methods}
\label{sec:exp_comparisons}

We use every pair of consecutive ground truth labels, from \textit{TrainEven}, to populate with segmentation labels the remaining 99 frames in between and evaluate on the center frame (from \textit{TrainOdd}), the one that is maximally distant from both manual labels and for which we have ground truth.


\begin{figure*}[t]
\begin{center}
\includegraphics[scale=0.19]{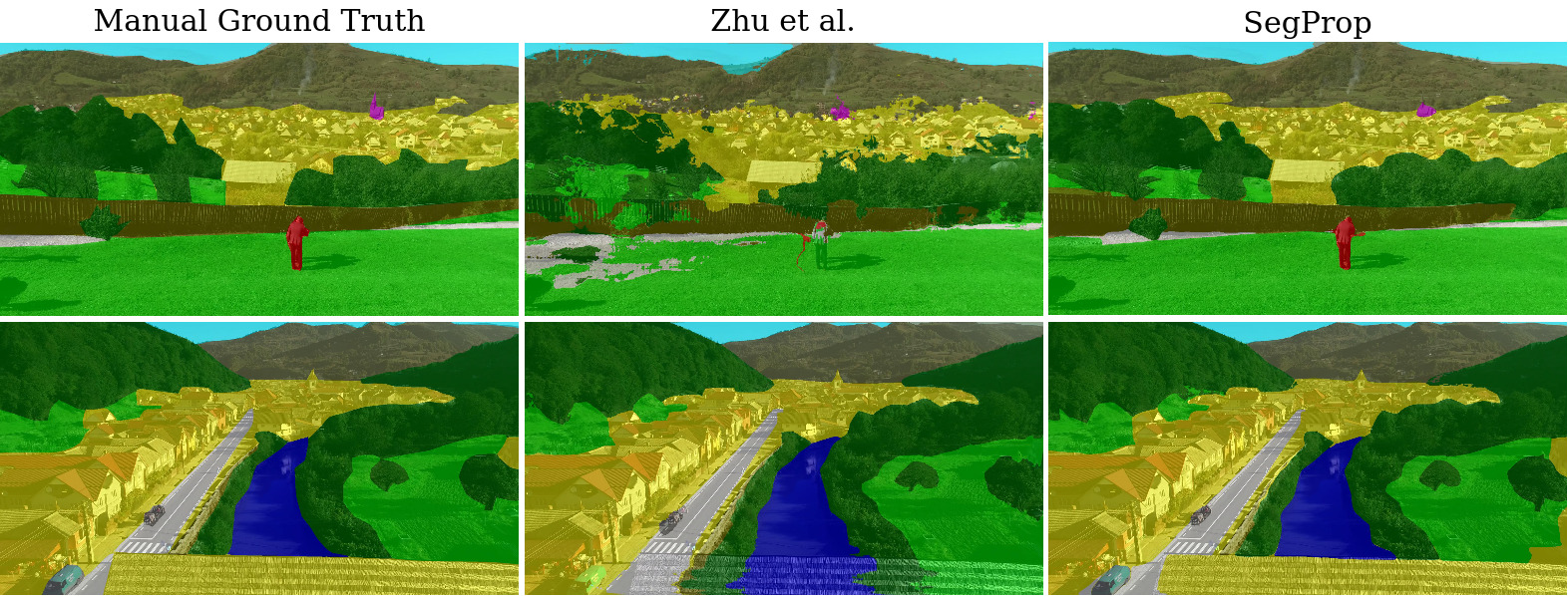}
   \caption{\label{fig:segprop_qual_results} Qualitative results of our label propagation method. Our iterative SegProp method provides labels that are less noisy and more consistent for larger propagation distances. Also, by looking both forward and backward in time we can better handle occlusion: this is easily visible on the second row in the bottom of the image where forward camera movement obscures a bridge.}
\end{center}
\end{figure*}

We compare our label propagation results with a state-of-the-art method recently proposed by Zhu et al.~\cite{zhu2019improving}. We use their method to similarly propagate ground truth labels. Since Zhu et al.'s method works with only one temporal direction at test time, we extract label estimations first forward and then backwards in time, up to a maximum of 50 frames, thus populating all 99 frames in between while keeping the propagation distance minimal. For a fair comparison, we used the same optical flow as~\cite{zhu2019improving}, namely FlowNet2~\cite{ilg2017flownet}. We test our SegProp (Sec. \ref{sec:segprop}) method against~\cite{zhu2019improving} and provide results in Table~\ref{table:label_propagation}. In Figure \ref{fig:segprop_qual_results} we also present some visual comparative results. We also show the effect of initialization on SegProp, when we start it with the solution from ~\cite{zhu2019improving} vs. initializing with Alg 1. SegProp improves in both cases and converges towards the same solution, but at different speeds. Finally, we apply our 3D filtering as a final step to further remove noisy labels and observe another final jump in F-measure.

\setlength{\tabcolsep}{4pt}
\begin{table}
\begin{center}
\caption{\label{table:label_propagation} Automatic label propagation comparisons. We measure mean F1-score and mean IOU over all classes. We present Zhu et al.~\cite{zhu2019improving} (which has only 1 iteration) vs. SegProp starting either from~\cite{zhu2019improving} or from Alg. 1 (our full SegProp Version). An interesting result, which confirms our theoretical expectation is that SegProp (Alg. 2) seems to converge to the same global relaxed solution, regardless of initialization (\cite{zhu2019improving} to Alg. 1), albeit with different convergence rates. The final output depends only on the structure of the graph (defined by the flow links) and the manually labeled frames, as expected.}
\begin{tabular}{|l|c|c|c|c|c|c|c|c|c|}\hline
\diagbox[width=10em]{Methods}{Iterations}
& & 1 & 2 & 3 & 4 & 5 & 6 & 7 & + Filt.\\ \hline\hline
Zhu et al.~\cite{zhu2019improving}&mF1& .846 & - & - & - & - & - & - & - \\
 &mIOU& .747 & - & - & - & - & - & - & - \\
 \hline
SegProp from ~\cite{zhu2019improving}&mF1& .846 & .874 & .877 & .885 & .888 & .891 & .893 & .896 \\ 
 &mIOU& .747 & .785 & .790 & .801 &.805 & .810 & .813 & .818 \\
 \hline
SegProp from Alg. 1 & mF1 & .884 & .894 & .896 & \textbf{.897} & \textbf{.897} & \textbf{.897} & \textbf{.897} & \textbf{.903} \\
 &mIOU& .801 & .817 & .819 & \textbf{.821} & \textbf{.821} & \textbf{.821} & \textbf{.821} & \textbf{.829} \\
 \hline
\end{tabular}
\end{center}
\end{table}
\setlength{\tabcolsep}{1.4pt}

We have tested different connectivity between frames for the iterative SegProp (Alg. 2)
and settled on the structure which connects the center frame $k$ to frames $[k-10, k-5, k, k+5, k+10]$. As computational costs increase both with the number of frames included in the set and with the distance between them, we find this to be a good compromise between width (inter-frame distance) and depth (number of iterations we can run). Note that $P_{k-10}$ at iteration 3 will have included propagated votes from $P_{k-20}$ on iteration 2, so width indirectly increases with iteration depth.

\subsection{Semi-supervised learning with automatically generated labels}
\label{sec:exp_semi_sup}

In order to assess the gain brought by the generated labels with SegProp, we train 3 different deep convolutional networks, two of which are widely adopted semantic segmentation models, namely Unet~\cite{ronneberger2015u} and DeepLabV3+~\cite{chen2018encoder}, and a model that has previously shown to yield good segmentation results on similar problems, in UAV flying scenarios~\cite{marcu2018safeuav}. Our approach, however, is agnostic of the chosen architecture and could work with any semantic segmentation method. We chose to train Unet since is the de facto standard for semantic segmentation networks and has been widely applied in many scenarios. Ultimately, our goal is to be able to deploy the model and use it on the UAV. Therefore, we trained two embeddable-hardware compatible deep convolutional networks, namely DeepLabv3+ with a MobileNetv2~\cite{sandler2018mobilenetv2} backbone and SafeUAV-Net Large. 

\textbf{Supervised baselines.} We also trained the same models only on the manually-labeled frames from \textit{TrainEven}. To compensate for the differences in terms of training volume, only for baselines, we apply data augmentation in the form of random rotations, color jittering and random flips, online, during training. 

\begin{table}[t]
\caption{\label{table:experimental_results} Quantitative results after training the neural networks on the generated labels. We report mean F-measure over all videos from the testing set, for each individual class: (1) - land, (2) - forest, (3) - residential, (4) - haystack, (5) - road, (6) - church, (7) - car, (8) - water, (9) - sky, (10) - hill, (11) - person, (12) - fence and the average over the all classes. The best results for each class and each trained model, are bolded. Results clearly show a significant performance boost over the baseline, when training with SegProp. (*) Due to space limitations we abbreviate SegProp with SP in the table.}


\label{table_results}
\begin{center}
\begin{tabular}{|c|c|c|c|c|c|c|c|c|c|c|c|c|c|c|}
\hline
Methods & SP* & (1) & (2) & (3) & (4) & (5) & (6) & (7) & (8) & (9) & (10) & (11) & (12) & All\\
\hline\hline
Unet & \xmark & .681 & .497 & .834 & .000 & .000 & .000 & .000 & .000 & \textbf{.967} & .000 & .000 & .000 & .248\\
\cite{ronneberger2015u} & \cmark & \textbf{.757} & \textbf{.544} & \textbf{.838} & .000 & \textbf{.556} & \textbf{.672} & .000 & .000 & .900 & \textbf{.454} & .000 & .000 & \textbf{.393}\\
\hline
DeepLab & \xmark & .500 & .416 & .745 & .000 & .220 & .073 & .000 & .000 & .909 & .242 & .000 & .000 & .259\\
v3+~\cite{chen2018encoder} & \cmark & \textbf{.570} & \textbf{.452} & \textbf{.776} & \textbf{.022} & \textbf{.369} & \textbf{.122} & \textbf{.007} & .000 & \textbf{.926} & \textbf{.272} & \textbf{.004} & \textbf{.043} & \textbf{.297}\\
\hline
SafeUAV & \xmark & .713 & .475 & .757 & .000 & .371 & .640 & .000 & .000 & .953 & \textbf{.260} & .000 & .003 & .348\\
Net~\cite{marcu2018safeuav} & \cmark & \textbf{.783} & \textbf{.488} & \textbf{.836} & \textbf{.364} & \textbf{.552} & \textbf{.748} & \textbf{.031} & \textbf{.428} & \textbf{.973} & .176 & \textbf{.481} & \textbf{.610} & \textbf{.515}\\

\hline
\end{tabular}
\end{center}
\end{table}

\textbf{Training details.} Models were trained using the same learning setup. Our deep learning framework of choice is Keras with a backend of Tensorflow. We use RMSprop optimizer with a learning rate starting from 1e-4 and decreasing it, no more than five times when optimization reaches a plateau. Training is done using the early stopping paradigm. We monitor the error on the validation set and suspend the training when the loss has not decayed for 10 epochs. The models were trained with RGB frames at a spatial resolution of $2048\times1080$px, rescaled from the original 4K resolution ($4096\times2160$px).

Quantitative results on the testing set are reported in Table \ref{table:experimental_results}. We compare our results with reference to the ground truth (manually given labels) from the testing set. The overall score was computed as mean F-measure over the whole classes. Some of the classes were not predicted at all in the supervised baseline case and were marked with .000. It is clear that all methods trained in a semi-supervised fashion
on the SegProp generated labels (marked with SP) strongly benefit from the label propagation procedure. The relative performance boost, compared to the supervised case, varies from 3.8\% for DeepLabv3+ with MobileNetv2 backbone, 14.5\% for Unet, and up to 16.7\% for SafeUAV-Net. The results also show that small classes experience a significant boost. The “secret” behind recovering classes that are completely lost lies in the label propagation algorithm that is able to add significantly more evidence for classes that are initially not well represented in the ground truth frames: appearing rarely, being very small or often occluded. The ambiguity for the land, forest and hill classes is reflected in our results. Well represented classes in the dataset such as residential areas and land, yield the best results. 

\begin{figure*}[t]
\begin{center}
\includegraphics[scale=0.2]{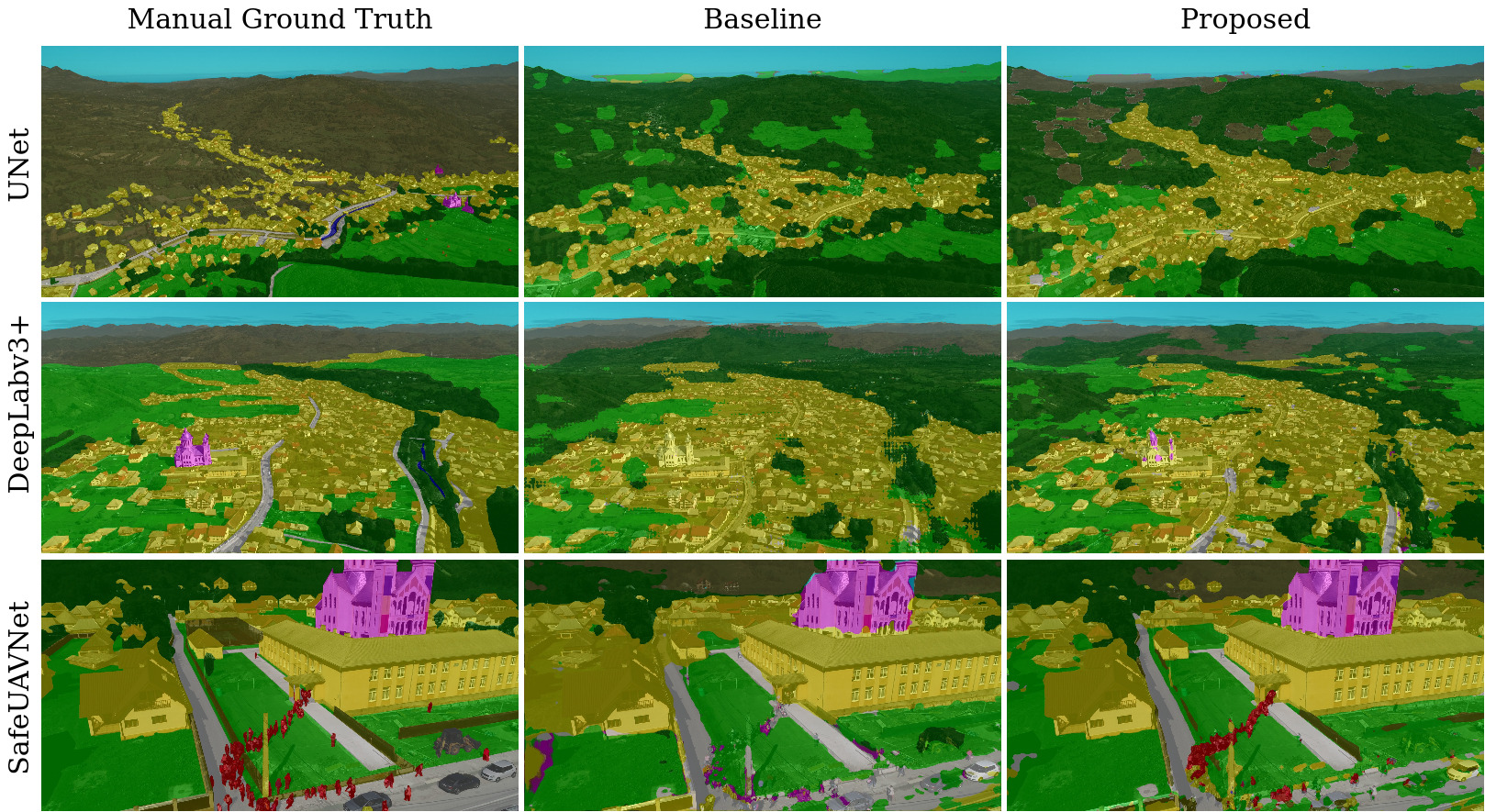}
   \caption{\label{fig:qual_results_training} Qualitative results on the testing set. The results show that our proposed method leads to significantly more accurate segmentation in the semi-supervised scenario rather than the supervised case. SegProp clearly benefits the smaller, not well represented, classes such as person (third row).}
\end{center}
\end{figure*}

Qualitative results on our testing set are shown in Figure \ref{fig:qual_results_training}. They exhibit good spatial coherency, even though the neural networks process each frame individually. The quality of segmentation is affected by sudden scene geometry changes, cases not well represented in the training videos and and motion blur.

\subsection{Ablation studies: the effect of the propagation module}
\label{sec:exp_ablation}

\begin{figure}[t]
\centering
\includegraphics[scale=0.2,keepaspectratio]{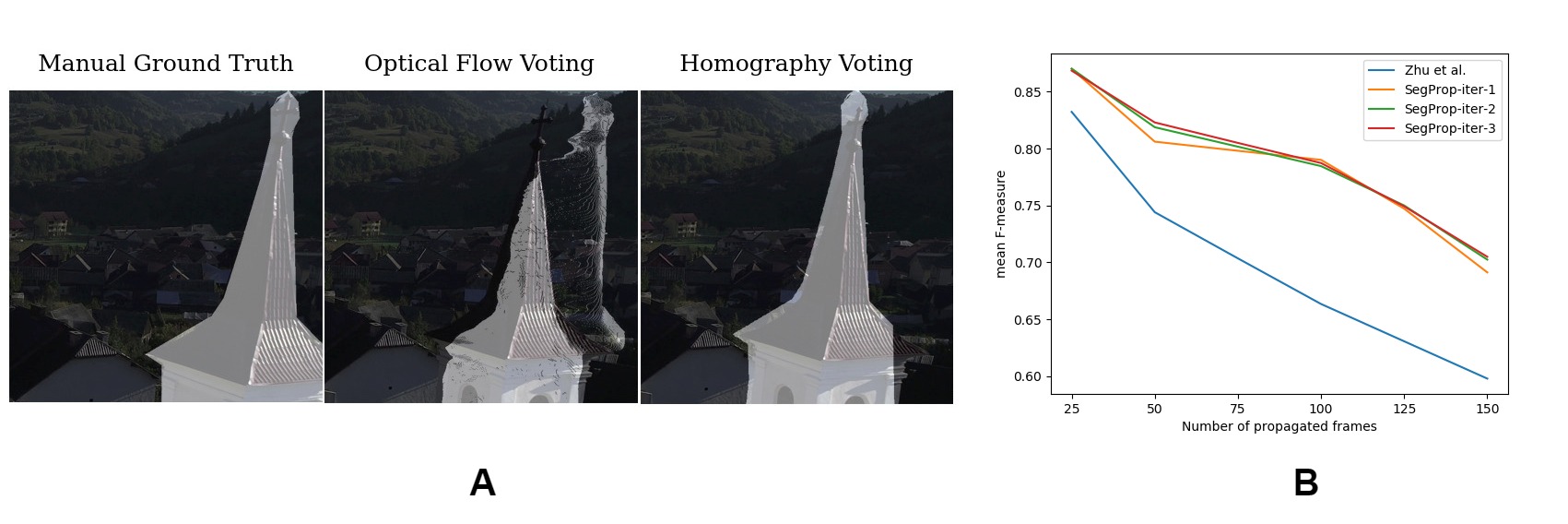}
\caption{\label{fig:homography_results} \textbf{A.} Label propagation example showing typical optical flow voting difficulties. From left to right: RGB frame with manual white label overlaid, flow-based voting, homography-based voting. \textbf{B.} The influence of increasingly larger temporal gaps between labeled frames over the segmentation performance (mean F-measure over all classes on a subset of videos labeled with a frequency of 25 frames).}
\end{figure}

\textbf{Homography propagation module.} Even state-of-the-art optical flow is prone to noise. In order to obtain more robust results, we test with the idea of incorporating geometric constraints to improve the class propagation. Thus, we compute two additional class voting maps coming from connected class regions in the labeled frames that are transformed with a homography and placed on the current frame of interest. The homography is estimated in a robust way, with RANSAC, using as correspondences the already computed flow maps between the labeled frames and the current one. Adding the homography based votes to the optical flow votes improves the results (Tab.~\ref{table:ablation_studies}) even from the first propagation iteration. Then, by applying our 3D filtering step on top, we further improve performance. While the homography based voting is clearly superior it is also much more computationally intensive, reason for which we did not include it in the other experiments presented in the paper. Note that voting propagation based on homography is particularly useful for edge preservation, where the CNN-based optical flow generally lacks precision (see Fig.~\ref{fig:homography_results} (A)).

\setlength{\tabcolsep}{4pt}
\begin{table}
\begin{center}
\caption{\label{table:ablation_studies} Ablation studies comparison. We run SegProp including other votes next to our optical flow based mappings, measuring mean F-measure over all classes. For the version with homography voting we also run the final filtering step. The bolded values are the best results.}
\begin{tabular}{|l|c|c|}
\hline
Method &  Iteration & Overall\\
\hline\hline
Zhu et al.~\cite{zhu2019improving} & 1 & .846\\
SegProp & 1 & .884 \\
SegProp + Zhu et al.~\cite{zhu2019improving} & 1 & .892 \\
SegProp + Homography & 1 & .894 \\
SegProp + Homography + Filtering & 1 & \textbf{.904} \\
\hline
\end{tabular}
\end{center}
\end{table}
\setlength{\tabcolsep}{1.4pt}


\textbf{Other vote propagation modules.} As mentioned in Sec.~\ref{sec:segprop} we could in principle use any label propagation method to bring in more votes.
Thus, in the same way we added homography voting to the initial optical flow ones, we also added two more class votes by using the method of 
Zhu et al.~\cite{zhu2019improving} to propagate class labels from the manually labeled frames to each unlabeled one. We weighted the votes with a validated parameter ($w = 0.25$) and observed another additional performance gain (see Tab.~\ref{table:ablation_studies}), even from the first iteration. Since it would have been computationally expensive to re-apply the method in~\cite{zhu2019improving} for voting, we have only tested with one iteration. Optical flow voting, while not the most accurate, remains very fast (computed only once at the start) and enables SegProp to achieve a significant boost over iterations.

\textbf{Influence of temporal propagation length.} We measured the degradation in performance as the propagation temporal length increases, from 25 frames to 150 frames and also compared with Zhu et al.~\cite{zhu2019improving} (Fig.~\ref{fig:homography_results} (B)). We performed the study on a subset of clips that are annotated every 25 frames ($1/3$ of the dataset), such that the evaluation can be done at every 25 frames as the propagation period increases. We measure mean F-measure over all classes from the selected videos. Note that our performance degrades slower than that of~\cite{zhu2019improving}.

\section{Conclusions}

We introduced SegProp, an efficient iterative label propagation algorithm for semi-supervised semantic segmentation in aerial videos. We also introduced Ruralscapes, the largest high resolution (4K) dataset for dense semantic segmentation in aerial videos from real UAV flights - which we make publicly available alongside a fast segmentation tool and our label propagation code\footnote{We make our code, dataset and annotation tool publicly available at:
\url{https://sites.google.com/site/aerialimageunderstanding/semantics-through-time-semi-supervised-segmentation-of-aerial-videos}}, in a bid to help aerial segmentation algorithms. We have demonstrated in extensive experiments that SegProp outperforms other published labeled propagation algorithms,
while also being able to work in conjunction with similar methods. Moreover, we have showed in semi-supervised learning experiments that deep neural networks for semantic image segmentation could extensively benefit (by up to $16\%$ increase in F-measure) from the added training labels using the proposed label propagation algorithm. SegProp is fast (it only needs to compute the optical flow once) and flexible, being able to integrate other label propagation methods. Furthermore, it has provable convergence and optimality properties. We believe that our work, introducing a well-needed dataset and algorithm, with strong experimental results, could bring a solid contribution to semantic segmentation in video and UAV research.

\noindent\textbf{Acknowledgements.} This work was funded by UEFISCDI, under Projects EEA-RO-2018-0496 and
PN-III-P1-1.2-PCCDI-2017-0734. We express our sincere thanks to Aurelian Marcu and The Center for Advanced Laser Technologies (CETAL) for providing access to  
their GPU computational resources.

\bibliographystyle{splncs}
\bibliography{egbib}

\title{Appendix}
\author{\textbf{Supplementary materials}}
\institute{}
\maketitle


More details and qualitative results are shown below to further demonstrate the effectiveness of our proposed method. We show the impact of homography on SegProp, qualitative results before and after training SegProp, the impact of adding only the iterative algorithm on top of other methods and timing details. Additional video content is included alongside this document.

\section{Discussion about convergence}

\begin{figure}
\begin{center}
\includegraphics[scale=0.5]{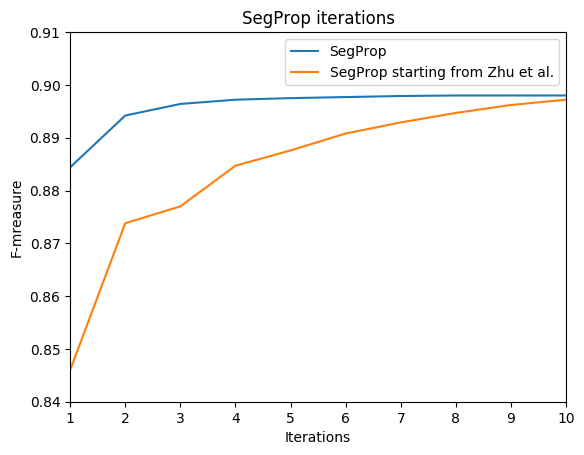}
   \caption{\label{fig:conv_plots} The iterative aspect of our algorithm manages to improve segmentations even more. Even though \cite{zhu2019improving} starts with poorer original segmentations, the iterations yield a better gain. However, having a good original segmentation helps as convergence should be achieved in fewer iterations.}
\end{center}
\vspace{-4mm}
\end{figure}

\begin{figure}
\begin{center}
\includegraphics[scale=0.2]{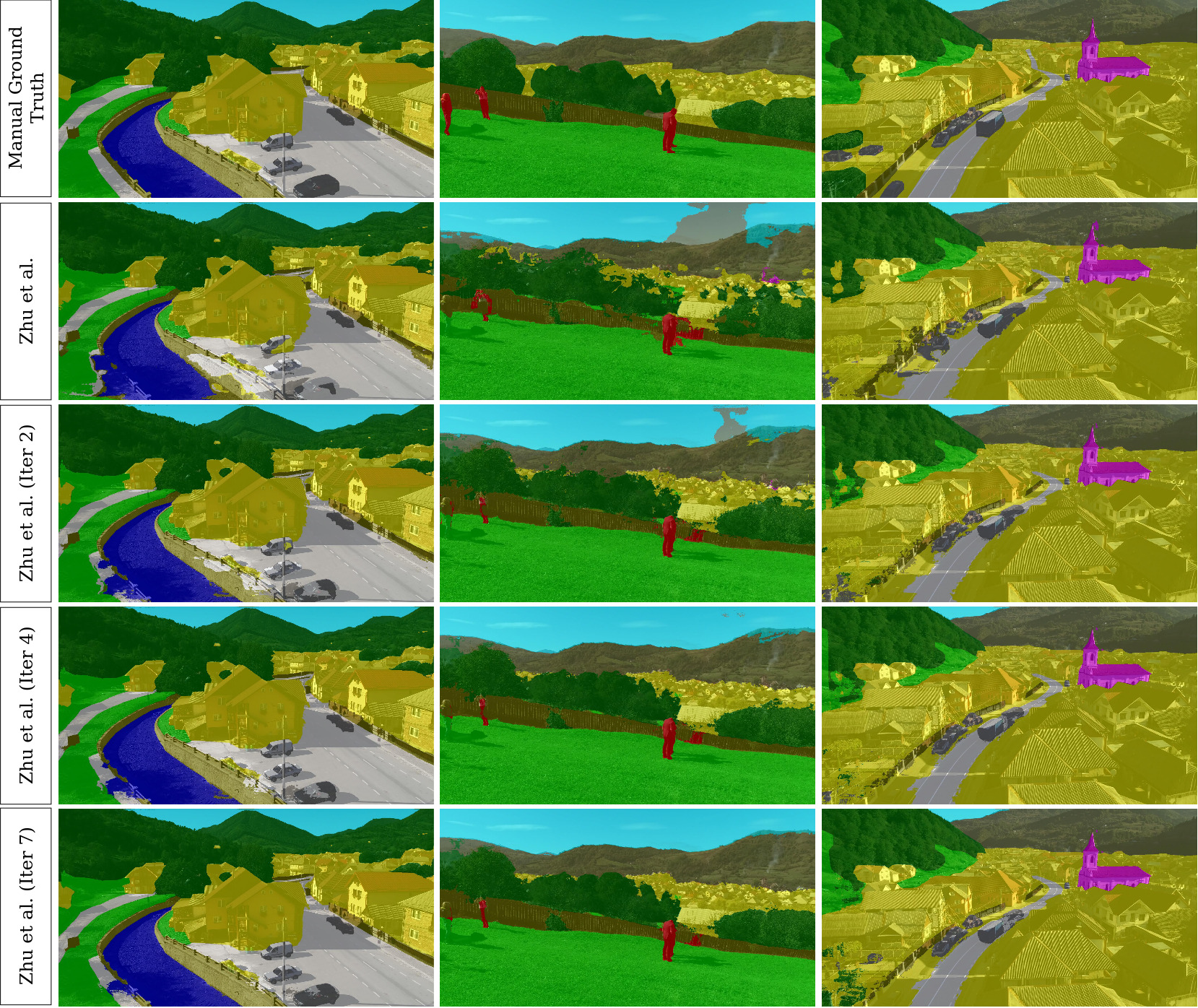}
   \caption{\label{fig:qual_segprop_nvidia} Qualitative results show the differences between competing methods, such as \cite{zhu2019improving} and our algorithms. The iterations contribute to the performance improvement, resulting in better details and better edge alignment, especially for smaller objects.}
\end{center}
\end{figure}

Here we present the numerical performance of SegProp in plot form (Figure \ref{fig:conv_plots}). The theoretical properties of our algorithm suggest convergence towards a singular value if enough iterations are computed, regardless of the starting point. What matters most is the static graph represented by optical flow and the ground truth data that is always forwarded unchanged on each iteration (see Section 2.1 in the paper). This progressive improvement can also be observed qualitatively in Figure \ref{fig:qual_segprop_nvidia}.
\vspace{-4mm}

\section{Adding Homography to SegProp}

Our method can support an arbitrary number of voting schemes. In the paper we present a homography based voting solution which we introduce after qualitatively assessing our initial results (Section 4.3). While Ruralscapes does not provide instance segmentations, we make the assumption that continuous labels are likely to correlate across a small enough interpolation distance $\Delta t$. Similarly, we assume that a mapping between two correlated regions can be approximated by a planar transformation for a sufficiently small $\Delta t$. We therefore identify connected components for each class map in $P_{i}$ and $P_{j}$ and project each such component onto $P_{k}$ by estimating a homography transformation between flow based correspondences -- we detail this method in Algorithm \ref{alg:homographyModule}. In practice, we use a least-median robust method ($LMEDS$) for estimating $H$ as a straight least squares derivation often fails for small objects due to the large number of outliers.

\begin{figure*}[t]
\begin{center}
\includegraphics[scale=0.19]{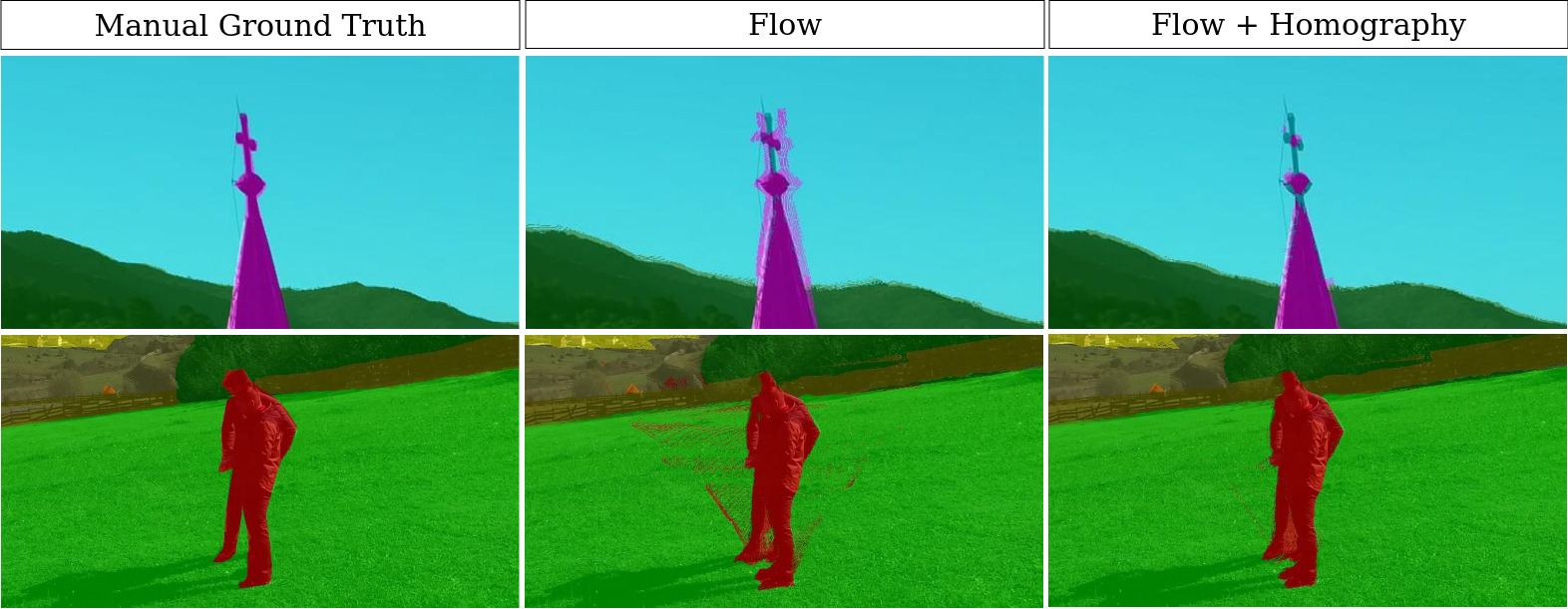}
   \caption{\label{fig:qual_homography} The influence of adding the homography based mapping to our voting pool. Camera movement causes objects on different planes of reference to move one against the other. Both optical flow errors and imperfect segmentations make label propagation difficult in this case, but a structure-preserving homography proves useful. We show image crops focusing on details.}
\end{center}
\vspace{-9mm}
\end{figure*}

Our intuition is that such a mapping will help the labeling of moving objects and will better preserve the segmentation edges. We support this idea with experimental results presented in the main paper (Table 3) and additional qualitative results shown in Figure \ref{fig:qual_homography}. Replacing our flow-based votes with the homography mappings instead of using them together was also tested, but the numerical results suffer as not all connected component transformations can be satisfyingly estimated.

\begin{algorithm}
\caption{Homography voting}
\label{alg:homographyModule}
\begin{algorithmic}
    \State 1) Generate an additional voting map by computing homography transformations between connected components $CC$ (connected regions with the same class label) from $P_i$ and their flow based correspondences on frame $k$:
    \For{each $CC$ in $P_i$}
        \For{$(x,y)$ in $CC$}
            \State $l_{i\rightarrow k}(x,y) = f_{i\rightarrow k}(x,y) + (x,y)$
        \EndFor
        \State $H_{i \rightarrow k} \leftarrow LMEDS(CC,L_{i\rightarrow k})$
        \For{$(x,y)$ in $CC$}
            \State $p^H_{i \rightarrow k}(x,y) = H_{i \rightarrow k}(p_i(x,y))$
        \EndFor
    \EndFor
    \State 2) Repeat the first step for $P_j$ and construct $P^H_{j \rightarrow k}$
    \State 3) Accumulate these two new votes with the first 4
\end{algorithmic}
\end{algorithm}

\section{SegProp - Discussion}

\textbf{Majority voting.} The final step of our algorithm is a simple majority vote - the class with the greatest cumulative score wins. However, it can happen that two or more classes share an equal maximum score - we estimate that approximately $0.12\%$ of total pixels suffer from this class uncertainty at decision time on the first pass of SegProp, and this number naturally decreases as more votes are counted in future iterations. In our current implementation there is no special handling of this state, the first class is selected by the $max()$ function from an arbitrarily ordered array. Future work could include better handling of this edge case, for example by counting neighbouring votes or considering a class priority list.

\begin{figure}
\begin{center}
\includegraphics[scale=0.19]{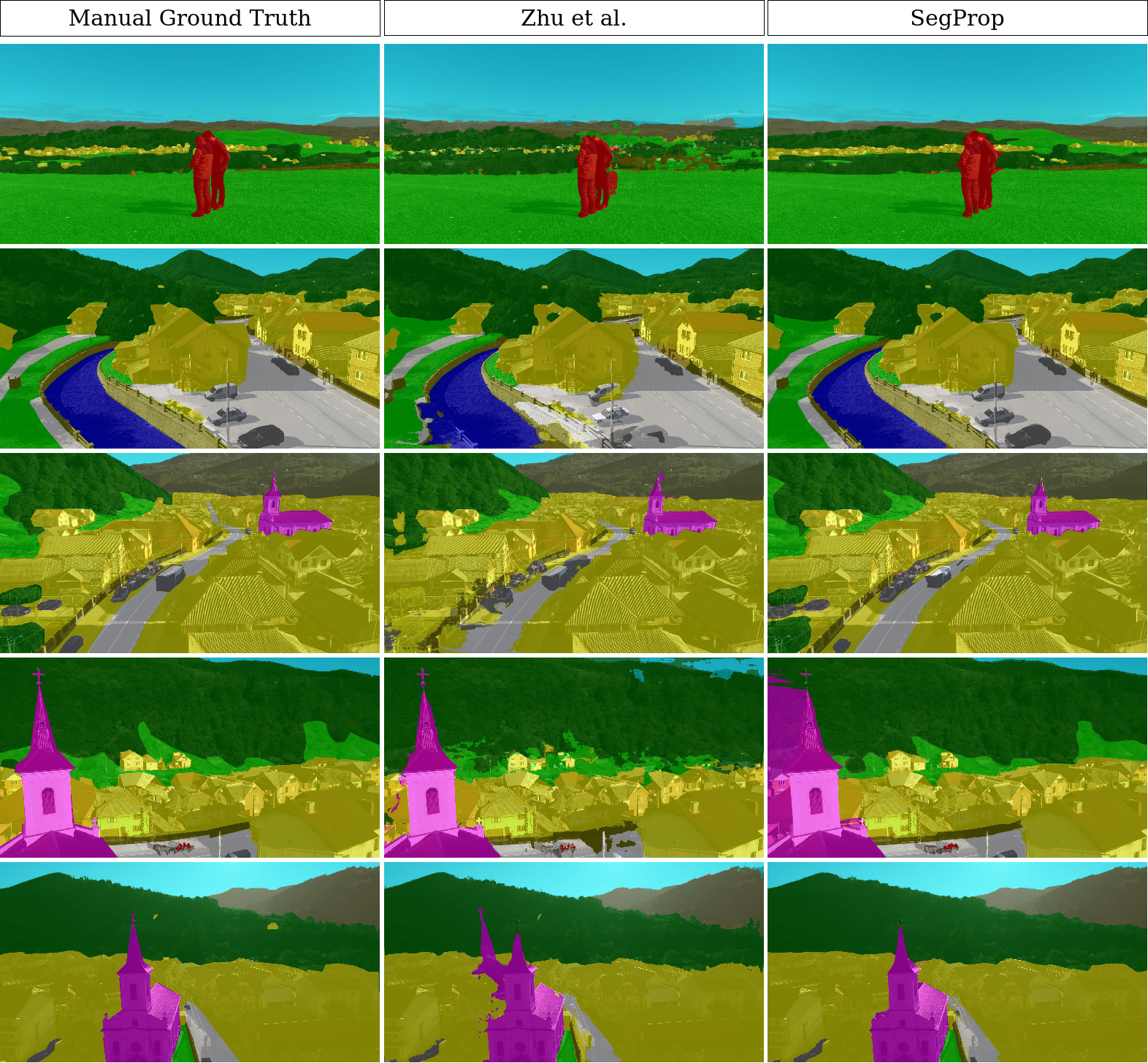}
   \caption{\label{fig:zhu_comp} SegProp compared to~\cite{zhu2019improving}. Better small object segmentation, better edges and more consistent detections are shown by SegProp. Heavily relying on optical flow and without a feedback mechanism, \cite{zhu2019improving} tends to result in inconsistent labels - see the fence, land, water and church areas from the images above.}
\end{center}
\vspace{-10mm}
\end{figure}

\textbf{Comparison with Zhu et al.\cite{zhu2019improving}.} While SegProp performs better both numerically and qualitatively for our use case, the method of Zhu et al. has at least one advantage over ours - the ability to propagate a single labeled frame, while SegProp requires a minimum of two. However, we achieve better results over larger time steps and on regions far away from the camera, at the cost of using an extra segmentation. 
Indeed, Zhu et al.~\cite{zhu2019improving} only use their method for relatively short distances of 1 to 5 frames, but for training purposes, segmentations might prove most useful when they are spaced further apart.
Another advantage of their method is the increased computational efficiency compared to our full iterative approach. However, we still achieve both better results (see Figure~\ref{fig:zhu_comp}) and faster running times using just one iteration (see Table~\ref{table:timing}).

\textbf{About the complexity of our task and approach.} Our aerial scenarios are in fact more difficult than many street-level car datasets. Ours has significant 6D pose changes (varied altitudes, viewpoints, rotations, 50kmph speed), varied and complex scenes, strong perspective effects and many different types of occlusions. The frame rate (50 fps) is high, but the number of propagated frames is also large. The actual propagation time is what matters most. Our algorithm is not simple in the way it uses iterative optical flow and homography voting, followed by 3D filtering. It is a form of spectral clustering, which is novel in video semantic segmentation literature. It is guaranteed to converge to the principal eigenvector of the space-time video graph, which ensures stability and global optimization under L2-norm constraints. That is the key reason why our SegProp, with different starting points, converges towards the same result (see Table~\ref{table:label_propagation}). 

\begin{figure}
\begin{center}
\includegraphics[scale=0.2]{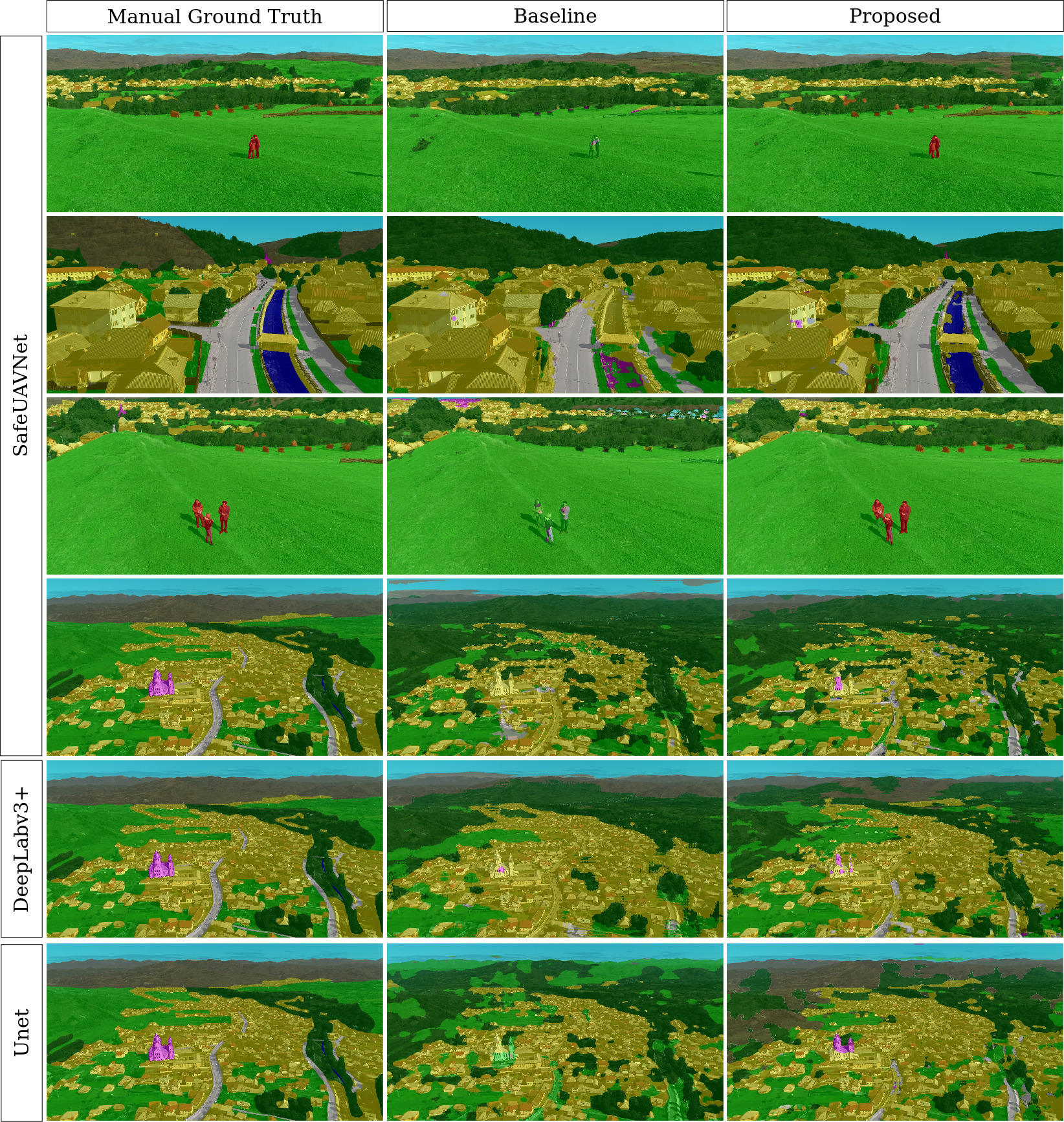}
   \caption{\label{fig:more_qual_results} Additional qualitative results on the testing set. Our method improves both the small and large object detection. For example, the humans in the third row are detected better, but also the skyline is more accurate (4th, 5th, 6th row). Even in uncertain label scenarios, such as the hill from the last row, our method yields a more plausible segmentation.}
\end{center}
\vspace{-9mm}
\end{figure}

\section{Qualitative SegProp results, after training}

Figure~\ref{fig:more_qual_results} presents more qualitative results for several state-of-the-art neural networks before and after training with our proposed method, SegProp. The first three rows show favourable results of our method compared to the baseline. The last three rows show the impact of the CNN choice in terms of performace - while the vanilla U-net and SafeUAVNet are similar, the former yields poorer results. DeepLabv3+ tends to fragment the labels, resulting in overall poorer segmentation.

\section{Timings}

Table~\ref{table:timing} presents the timing requirements of our method. While a single iteration is faster than~\cite{zhu2019improving}, generating almost one frame per second at $2048\times1080$px, adding homography or iterations increases the computational cost. Nevertheless, there is a linear cost associated with the iterations - the algorithm can be stopped when timing constraints are reached.
\vspace{-3mm}
\begin{table}
\begin{center}
\caption{\label{table:timing}Timing results for one frame. The numbers below are computed for the rescaled images ($2048\times1080$px).}
\begin{tabular}{|l|c|}
\hline
Method  & Runtime (seconds)\\
\hline\hline
Zhu et al.~\cite{zhu2019improving} &  1.74 \\
SegProp Iteration 1 & 1.12 \\
SegProp Iteration 1 + Homography & 12 \\
SegProp Iteration $N$, with $N > 1$ & 5.14 \\
\hline
\end{tabular}
\end{center}
\vspace{-7mm}
\end{table}

\end{document}